# Machine learning for cloud resources management - An overview

Viktoria N. Tsakalidou, Pavlina Mitsou, George A. Papakostas

**Abstract** Nowadays, an important topic that is considered a lot is how to integrate Machine Learning(ML) to cloud resources management. In this study, our goal is to explore the most important cloud resources management issues that have been combined with ML and which present many promising results. To accomplish this, we used chronological charts based on some keywords that we considered important and tried to answer the question: is ML suitable for resources management problems in the cloud ? Furthermore, a short discussion takes place on the data that are available and the open challenges on it. A big collection of researches is used to make sensible comparisons between the ML techniques that are used in the different kind of cloud resources management fields and we propose the most suitable ML model for each field.

## 1 Introduction

In July 2020, in a demographic research conducted by Statista, approximately 4.57 billion users were active on the Internet, encompassing 59% of the global population [22]. This huge amount of users shows the demanding need for resilient, secure and easily configurable web applications. Before the cloud computing era, from the enterprise perspective, the cost of the maintenance of big data centers and bootstrap-

Viktoria N. Tsakalidou
HUman-MAchines Interaction Laboratory (HUMAIN-Lab), Department of Computer Science, International Hellenic University, Kavala, 65404 Greece , e-mail: vitaaka@cs.ihu.gr

Pavlina Mitsou
HUman-MAchines Interaction Laboratory (HUMAIN-Lab), Department of Computer Science, International Hellenic University, Kavala, 65404 Greece , e-mail: pamitso@cs.ihu.gr

George A. Papakostas
HUman-MAchines Interaction Laboratory (HUMAIN-Lab), Department of Computer Science, International Hellenic University, Kavala, 65404 Greece , e-mail: gpapak@cs.ihu.gr

ping a company was too high. *Cloud Computing* is the provision of virtual resources via the Internet (e.g. servers, apps, etc.), from central systems located away from the end users, which serves them by automating processes, providing convenience, flexibility of connection[18] as well as a nice pay-as-you-go payment plan to the company. In the 1950s first appeared in educational institutes and companies, and its use was made by servers with large computing and storage capabilities[18].

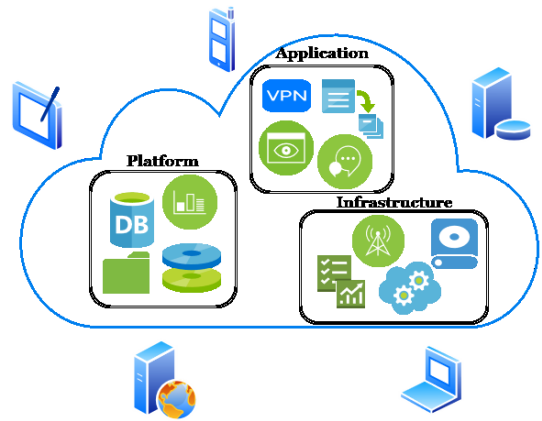

**Fig. 1** Basic cloud service categories.

As Fig. 1 depicts three are the basic cloud service categories:

- *Software-as-a-Service (SaaS)*: There are already multiple applications that they are provided as SaaS like CRM, ERP, Word etc. So the main concept is to not install the software on the customer's computer or servers but to access them via internet. In that way the release process, and the bug fixes will be easier [18], [17].
- *Platform-as-a-Service (PaaS)*: There is also a similar concept as SaaS but in PaaS as an alternative of having to buy - pay for software licenses for platforms such as operating systems, databases and intermediate software, the customer can do so using the platform and tools [18], [17].
- *Infrastructure-as-a-Service (IaaS)*: The IaaS is even in a bigger scale and abstraction. So instead of owning and maintaining your own hardware devices in your data center, you can access them via the internet by logging in with official credentials and they can be distributed all around the planet. These hardware devices can be networking related devices, VMs, storage devices etc. [18], [17].

A commitment between a service provider and a client called Service-level Agreement (SLA). Wieder et al. [9] mentioned that some aspects of service quality, responsibilities, availability, have to be agreed between the service user and service provider. SLA usually contains a lot of components, from a definition of services to the termination of the agreement. Also, SLA defined at different levels:

- *Multilevel SLA*: is a partition into different levels, each addressing a different set of customers for the same services, in the same SLA. The levels are: Corporate-level SLA, Customer-level SLA and Service-level SLA.
- *Customer-based SLA*: An individual agreement in a customer group that contains all the services they use.
- *Service-based SLA*: In a common deal with all customers using the services being delivered by the service provider.

The *Quality of Service* (QoS) is the measurement or description of the overall performance of a service, such as a Cloud Computing service, or a computer network, particularly the performance seen by the users of the network. Many components of network service are often considered to measure service quality such as availability, transmission delay, packet loss, etc.

Arthur Samuel, in 1959, coined the term *Machine Learning* (ML). He was a pioneer in the field of computer gaming and AI [21],[20]. Machine Learning is the ability of the machines to learn from data. The ML makes computers to discover how to do tasks without being programmed explicitly. It steadily makes its unique way in different kind of areas in enterprise applications such as fraud detection, business intelligence and customer support. Microsoft, Amazon, and Google made significant investments in Machine Learning and Artificial Intelligence (AI), by launching new services to carry out important reorganizations which will strategically place Artificial Intelligence in their organizational structures in the last years. Even Google's CEO, Sundar Pichai, mentioned that Google is shifting to an " AI-first" world. Some authors said that the hardware is impacted by ML workloads. In order to train a model to understand speech or recognize a pattern requires major parallel computing resources that could take days with the traditional computers based on CPU. We can say that, powerful Graphics Processing Units (GPUs) is the first choice for the processing unit in many Machine Learning and Al workloads due to significantly reduced processing time.

This paper aims to present the applications of Machine Learning in the cloud resources management fields and the challenges that still remain open. There is an extensive investigation among the most popular topics in cloud resources management on the effectiveness of ML models and the metrics that are used to measure the performance. Also, a discussion on the popularity of this topic in the research community until now is included.

The structure of this paper is as follows. On each of the subsections of Section 2 is presented the related work that has been done in each cloud resources management field that ML models can be applied. Some comparisons and comments between the ML models and the data are taken place on it too. In Section 3 there is a discussion around the dataset benchmarks that have been used in the research until now and the way that are used. There is an extensive discussion in the open challenges that are still relevant and important in Section 4 and the future work that can be conducted around this topic. Eventually, Section 5 concludes the paper.

## 2 Machine Learning in Cloud Computing

ML and cloud computing have been trends and extensively used all around the tech world. Cloud provides ML solutions throughout it but there is also a tendency in the research that explores how the cloud resources management will be benefit from ML. From 2009 till 2019 there is a constant increase in the numbers of papers that are related to the usage of ML in cloud computing as it shows in Fig. 2. These results are based on the 300 most related articles in Google Scholar. The keywords that were used to extract the results were "Cloud Resources Management" and "Machine Learning". Figure 3 presents the number of articles on different kinds of cloud resources management fields. For each distinct field, the number of papers that are used for the graphs was figured by checking that the field was part of the paper's title and it was explicitly for that. The articles that we decided to focus were the one that the number of citations was high and the paper title was more relevant to the field. Although a valid question is "Machine learning suitable for cloud resources management problems?".

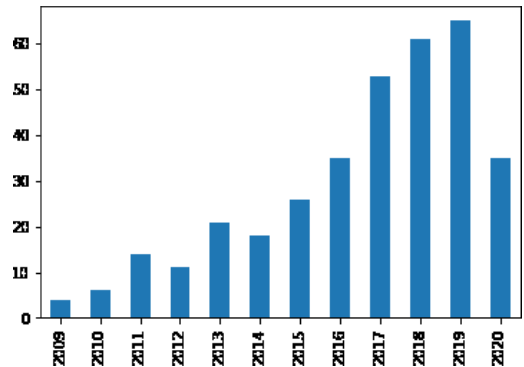

**Fig. 2** Scientific articles published for "Machine Learning for Cloud Resources Management" per year.

### 2.1 Resource provisioning

*Resource provisioning* has as parts the deployment, selection and management of hardware and software resources in run time for ensuring guaranteed performance for applications. Cloud providers have two provisioning options. First is the *short-term on-demand* and the second is *long-term reservation plans*. [18] There are important and challenging problems in the large-scale distributed systems that Resource provisioning is related to. Some of the resource delivery techniques used must meet service quality (QoS) parameters, such as security, availability, reliability, etc., avoiding Service Level Agreement (SLA) violations and managing long-term reservation plans.

Hanieh Alipour and Yan Liu proposed a method to make predictions and learn workload patterns with real time load on a microservice architecture that used ML models.[3]. They used AWS, to demonstrate the microservice architecture and 2 ML algorithms of logistic regression and multiclass classification. The ML models they used were the linear regression and the multinomial logistic regression. They integrated the real time prediction to the auto-scaling configuration to remove or add resources.

Sadeka Islam et al. [8] developed prediction-based resource measurement and provisioning strategies using Neural Networks(NN) and Linear Regression models to satisfy upcoming resource demands. They proposed prediction techniques in the context of the dataset obtained by using TPC-W, a benchmark that is well-established for e-commerce applications. The prediction accuracy of the proposed learning models is promising and effective.

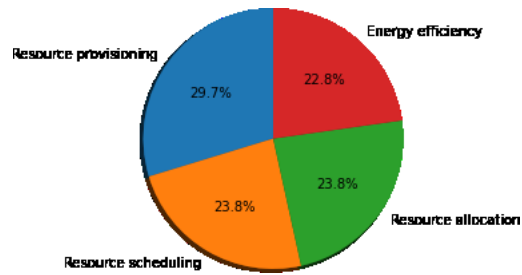

**Fig. 3** Scientific articles published per section in total that involve ML solutions.

Cloud client prediction models using Machine Learning is also extensively researched by A. Bankole and S. Ajila [11] through the TPC-W benchmark web application in Resource provisioning by using three different Machine Learning techniques. Among Linear Regression (LR), Neural Networks (NN) and Support Vector, it seemed SVM was the best prediction model by focusing on CPU utilization and SLA (throughput and response time) in a 9-12 minutes window.

Based on the prediction of the processing load that a distributed server can have and the estimation of the appropriate number of resources that must be provisioned, R.Moreno-Vozmediano et al. created an autoscaling predictive framework using ML methods for queuing theory and time-forecasting [10]. For the prediction the SVM regression model was used, which appears to be ideal for the data input with linear and non-linear patterns.

According to the above researches, it seems that Resource provisioning and ML are used in combination in order to predict the upcoming resource demands. The most common ML techniques, which their results are promising and efficient, are Linear Regression (LR), Neural Networks (NN), and Support Vector Machine (SVM). Although based on Table 1 SVM appears to outperform the rest ML techniques in a variety of metrics and datasets.

**Table 1** Related work for *Resource provisioning*

| Ref. | Year | Prediction Model | Metric | Performance | Dataset |
|------|------|------------------|--------|-------------|---------|
| [10] | 2019 | SVM - based forecasting models | SLA Violation | Outperform the basic forecasting models | Real Dataset from University of Madrid |
| [3] | 2017 | Logistic and Linear Regression | CPU usage | RMSE = 1.9426 | Netflix Data Benchmark [29] |
| [24] | 2014 | Linear Regression and SVM | Amount of profit broker earns per hour in $ | SVM better that LR by 2% - 3% | Data collected from the application |
| [11] | 2013 | NN, SVM and Linear Regression | CPU and SLA | SVM outperforms NN and LR in the MAPE and PRED metrics | TCP-W Benchmark [28] |
| [8] | 2012 | NN and Linear Regression | Response time | RMSE of NN = 9.6 RMSE of LR = 10.41 | TPC-Benchmark |

## 2.2 Resource scheduling

*Resource scheduling* allow us to efficiently schedule and automate any virtual job. It is a multi-dimensional work in cloud as it is depending on QoS of cloud apps. Cloud is by nature a diverse environment, so it encounters a variety of resources and uncertainty on the allocation of resources, which makes it impossible to be solved with existing the resource allocation policies. Until today research community struggles to choose an efficient algorithm for the resource scheduling on selected workload from the existing literature.

Michal Wilk et al. [16] proposed a system with reservation scheme adaptation that used ML. It is considered important in the automated scaling the problem to define a reservation program, which allows effective resource scheduling that can be used. Based on the above issue they found a solution that allows updating a booking program originally drawn up by an administration. This will allow reservation plans to be adjusted one or more weeks in advance. In this way, the administrator has the opportunity to identify potential problems with the under-provisioning or over-provisioning on the resources. This could help to prevent the server of unnecessary costs and overloads later on. That resolution tested on OpenStack and it was used real traffic data from Wikipedia. First they used Neural Networks because were applied successfully to assign allocation adaptation in Mobile Cloud. Afterwards, they compared the quality and the results among linear regression and decision trees techniques.

Authors compared the performance of the Neural Networks with the other three ML algorithms which were the RepTree , the linear regression, and the M5P. In order to find and compare the results, they summed the Q values of all 365 days. They understood that, the RepTree algorithm was learned 7% faster than Neural Networks, but the latter model ultimately yielded better predictions.

They authors claimed that is closed-loop solution. So the big benefit is that it can be verified in advanced the updated plan in order to identify potential issues in short and longer periods. Also, proposed, that before the deployment, we can take the advantage of the historical application's data for the parameter setting and the initial training of the machine learning models. They believed, that from the experimental results they had, this solution improves the use of cloud resources. They claim that the proposed method can be applied to other resources and it is more general. [16].

Renyu Yang et al. [7] proposed a general based solution through data driven profiling, problem formalizing, modeling with supervised, and gave a complete architecture. They believed that could achieve optimization in architectural level and improved efficiency with an ML based method. As result they said that they consider important to pair ML algorithms and systems on edge and cloud devices.

Sukhpal Singh and Inderveer Chana [25] mentioned that in literature, cost, time constraints and energy are some of the efficient resource scheduling policies that have been reported. They proposed a framework which was efficient on cloud management to identify and analyze the workload. K-means was used to cluster them and as weights was the QoS requirements. The experimental results were collected by CloudSim. They proposed that, this framework showed improved performance on cost, execution time, energy consumption unlike existing scheduling algorithms.

The main focus of resource scheduling is how to create a cost-efficient reservation plan. For this mission different kind of ML models have been used in the researches based on Table 2 and QoS is a metric that there is a lot of attention on it.

But still, the development of such mechanisms with this intelligent scheduling abilities becomes subsequently more challenging and complex.

### 2.3  Resource allocation

*Resource allocation* includes the assignment of available resources to the Cloud apps that needs them over the internet. Resource allocation needs to be managed precisely. If there will not be, it will starve from services. That problem can be solved with the resource provisioning which allows the management of every module independently by the service providers.

**Table 2** Related work for *Resource scheduling*

| Ref. | Year | Prediction Model | Metric | Performance | Dataset |
|------|------|------------------|--------|-------------|---------|
| [16] | 2019 | NN, Linear Regression, RepTree and M5P | CPU usage | RepTree algorithm is 7% better than NN | Wikipedia server data and Grafana Wikimedia |
| [7] | 2018 | Decition Tree, NN | QoS and Accuracy | prediction accuracy of node performance 92.86% | Open Cloud cluster at Carnegie Mellon University |
| [25] | 2015 | K-means | CWMF, QoS and Energy comsumption | Energy consumption was reduced from 9.99% on a hundred resources and 11.02% on the tripled amount of resources. | Standard Performance Evaluation Corporation (SPEC) |

Ning Liu et al. [15] proposed a framework that claims to solve the overall power management and resource allocation issues in cloud. It includes a tier for virtual machine global resource allocation to servers and for distributed energy management, a local tier. Except the improved scalability and the reduced action/state space dimensions, by performing local power management on the servers, it enhances the parallelism and it reduces the complexity computationally. In order the global tier problem to be solved a DRL technique is adopted. For the acceleration, a structure which has unique weight sharing was used. LSTM workload predictor helped power manager, without a model, based on Reinforcement Learning (RL) to determine a suitable action for the servers.

For experimental purposes actual Google cluster traces were used. This framework has this ability to save 53.97% energy consumption, with 95k job requests in just 30-server cluster, [14]. In the identical case, this framework gave the average per-job latency saving with the identical energy usage up to 16.16% , and the average power/energy saving with the identical latency up to 16.20%. ccording to their experiments they found that the proposed hierarchical framework significantly saved the energy consumption compared to the baseline with similar average latency. Moreover, in a server cluster, it could achieve the best trade-off among latency and power consumption[15].

Jixian Zhang et al. [6], proposed a model to analyze the resource allocation, which is a multi dimensional problem. They proposed two algorithms for the resource allocation prediction by using logistic and linear Rrgressions. They found out that the algorithm can provide a practical solution which is close to the optimal based on resource utilization and allocation accuracy.

According to Table 3, despite the fact that resource allocation is considered an NP-hard problem, it is possible to do resource allocation predictions by using Linear Regression and the results can be really close to the optimal.

**Table 3** Related work for *Resource allocation*

| Ref. | Year | Prediction Model | Metric | Performance | Dataset |
|------|------|------------------|--------|-------------|---------|
| [6] | 2018 | Logistic and Linear Regression | Response time | 98% prediction accuracy | DAS-2 dataset [30] |
| [15] | 2017 | Deep Reinforcement Learning | power consumption | reduced energy by 16.20% | Google cluster-usage traces [27] |
| [26] | 2013 | SVM and KNN | CPU utilization | SVM outperforms KNN | MediaPaaS platform |

## 2.4 Energy efficiency

Has become a major concern in data centers due to the environmental impact, costs and operational expenses that imparts as problems. The infrastructure still needs planing for the resources in order to address the energy wastes although it has been reduced a lot lately due to the utilization of best-practice technologies. The hardware devices and technologies have been improved based on energy consumption a lot but still some concerns exist to their usage. [13].

Heath et al. [12] analyzed how to distribute the requests from clients to different servers in a cluster with heterogeneous form in which an optimal balance among energy savings and throughput will be considered. They designed and developed a cluster which can be configured itself in order metrics such as energy consumption to be optimised. Analytical models were used the distribution of requests and the resource utilization. Simulated annealing on request distribution was used to find the minimum ratio on power to throughput. This particular method appeared to have energy consumption savings more than 40%.

John J. Prevost et al. [5] presented a framework by combining stochastic state transition models and load demand prediction. It was used NN and linear regression models. They claimed that their model will lead to optimal cloud resource allocation by minimizing the energy consumed while maintaining required performance levels. Finally, they found out that from the simulations that linear predictor and NN seemed to have promising predictions for the future network loads. But the linear predictor eventually provided the most accurate results.

Truong Vinh Truong Duy et al. [4] introduced a scheduling algorithm which also characterized as a green algorithm that uses a predictor for power savings based on NN in cloud. They have conducted various experiments with several simulation configurations. They concluded that the optimal configuration to assure service level is the prediction in addition 20% more servers. They proposed that the power consumption savings for PP20 mode can achieve up to 46.3%.

In Energy efficiency, the results are promising since helps to reduce energy almost by half by optimally configuring CPU frequency and propose the most energy saving configurations. Neural Networks is an efficient ML model in multiple datasets but the Linear Predictor is more accurate as it can be seen from Table 4.

**Table 4** Related work for *Energy efficiency*

| Ref. | Year | Prediction Model | Metric | Performance | Dataset |
|------|------|------------------|--------|-------------|---------|
| [5] | 2011 | NN and Linear Predictor | RMSE | RMSE for the LP .25 and for NN over 8.00 | NASA and EPA |
| [4] | 2010 | NN | energy consumption | Rate power reduced by 46.7% on NASA | CISRegistry |
| [12] | 2005 | Model Adaptive | Less Energy | Their server conserves 45% more energy | Microbenhmark |

## 3 Datasets

In the ML, the datasets are an integral part. Cloud every day produces tones of data from the resource management processes but it seems that there are not available open data to be used in the research. In the vast majority of papers that we analyzed we observed that tools generate mostly the datasets [29] [28] or they are internal datasets that are used. There are not so many open datasets for analysis purposes and that have been used in common research works in order to be able to have better performance metrics and comparisons. The open data mostly contains information which are traces and logs on the machine events, the requested resources, scheduling information, etc. Usually, these open datasets are provided by files but also some others by SQL scripts, BigQuery tables which facilitates the analysis.

## 4 Discussion

Machine learning for cloud resources management is an issue that has been discussed for the last 10-15 years. As shown in Fig. 2, from 2009 to 2019, there is a steady increase in the number of papers related to the usage of ML in cloud computing, but in 2020 this steady increase has fallen. We believe that it may be correlated with the phenomenon of Covid-19 pandemic because it is something demanding and there are a lot of data available on it. So most researchers would like to deal with this issue as it is considered to be a pandemic and it has influenced all of the people's lives this period. Also, due to cloud resources management which requires complicated decisions and policies for multiple objective optimization and therefore the effective resource management is challenging because of the scale of the cloud infrastructure. Based on the papers that are written in the field, seems that the large population of users combined with the unpredictable interactions of the system and the ML difficulties to integrate makes it challenging to solve such problems. An open challenge is also the data that is not easily accessed and available for the researchers to use them. It is observed that Machine Learning significantly helps to reduce energy consumption almost by half, by optimally configuring CPU frequency and providing the most energy saving configurations. Also, the results from Resource allocation predictions by using Linear Regression can be really close to the optimal. It can be concluded from most of the sections that Machine Learning is a really suitable solution in all of the cloud resources management fields. Mostly it helps with the prediction of future resource demands.

## 5 Conclusion

In this paper, we have tried to include the most important cloud resources management issues that are handled by the Machine Learning paradigm showing promising results. In particular, we observed from the conducted literature analysis that Machine Learning is the most suitable solution in cloud resources management issues. The most popular models, that are used, are the Neural Networks (NN), the Linear Regression (LR), and the Support Vector Machine (SVM). We concluded that the ML can provide feasible solutions that are very close to the optimal ones based on allocation accuracy, energy saving and resource utilization. Furthermore, we realized that there is a serious problem with the data that are available. Researchers have generated a variety of in-house data, but there are not available open data in order to be used in other researches and thus we cannot have sensible comparisons on them. The evaluation of several ML oriented methodologies for solving the cloud resources management issues and their comparison with popular non ML-based techniques on the same data constitute potential directions for future research.

**Acknowledgements** This work was supported by the MPhil program "Advanced Technologies in Informatics and Computers", hosted by the Department of Computer Science, International Hellenic University, Greece.